\documentclass[runningheads]{llncs}

\usepackage{eccv}

\usepackage{eccvabbrv}

\usepackage{graphicx}
\usepackage{booktabs}

\usepackage{hyperref}

\hypersetup{
    colorlinks=true,       
    urlcolor=magenta       
}

\usepackage{orcidlink}

\begin{document}

\title{DiVR: incorporating context from diverse VR scenes for human trajectory prediction}

\titlerunning{DiVR: Incorporating context from VR for Human Trajectory Prediction}

\author{Franz Franco Gallo\inst{1}\orcidlink{0009-0007-3859-7331} \and
Hui-Yin Wu\inst{1}\orcidlink{0000-0001-7315-210X} \and
Lucile Sassatelli\inst{2,3}\orcidlink{0000-0003-1232-1787}}

\authorrunning{F.Franco et al.}

\institute{Université Côte d'Azur, Inria, Sophia-Antipolis, France \and
Université Côte d'Azur, CNRS, I3S, Sophia-Antipolis, France
 \and Institut Universitaire de France, Paris, France\\
\email{franz.franco-gallo@inria.fr}}

\maketitle

\begin{abstract}

Virtual environments provide a rich and controlled setting for collecting detailed data on human behavior, offering unique opportunities for predicting human trajectories in dynamic scenes. However, most existing approaches have overlooked the potential of these environments, focusing instead on static contexts without considering user-specific factors. Employing the CREATTIVE3D dataset, our work models trajectories recorded in virtual reality (VR) scenes for diverse situations including road-crossing tasks with user interactions and simulated visual impairments. We propose Diverse Context VR Human Motion Prediction (DiVR), a cross-modal transformer based on the Perceiver architecture that integrates both static and dynamic scene context using a heterogeneous graph convolution network. We conduct extensive experiments comparing DiVR against existing architectures including MLP, LSTM, and transformers with gaze and point cloud context. Additionally, we also stress test our model's generalizability across different users, tasks, and scenes. Results show that DiVR achieves higher accuracy and adaptability compared to other models and to static graphs. This work highlights the advantages of using VR datasets for context-aware human trajectory modeling, with potential applications in enhancing user experiences in the metaverse. Our source code is publicly available at \href{https://gitlab.inria.fr/ffrancog/creattive3d-divr-model}{https://gitlab.inria.fr/ffrancog/creattive3d-divr-model}.

  \keywords{Virtual Reality \and Scene understanding \and Heterogeneous graphs \and Cross-modal}
  
\end{abstract}

\section{Introduction}
\label{sec:intro}

Capturing dynamic interactions between individuals and their environments is crucial for human motion prediction, and including visual scene context enhances the accuracy of these predictions. However, existing approaches primarily rely on frame-by-frame video analysis, and have difficulty to adequately account for these complex interactions \cite{smith2020limitations}. Scene point clouds generated from 3D sensors provide spatial data but can't capture temporal changes nor the human intentions \cite{zheng2022gimo,jones2021challenges}. These data capture and representation methods, though useful in controlled and simple scenarios, fall short in offering a nuanced understanding of individual intentions and interactions within dynamic environments. This limitation significantly impacts prediction accuracy, particularly in complex navigation scenarios or for individuals with invisible conditions such as visual impairments. 

\setcounter{footnote}{0} 

VR provides a secure and controlled media to replicate real-world scenarios to study human behavior. The CREATTIVE3D dataset \cite{wu2023exploring} \footnote{\href{https://zenodo.org/doi/10.5281/zenodo.8269108}{https://zenodo.org/doi/10.5281/zenodo.8269108}}, integral to this research, offers ontology-based VR environments, with annotations of 3D scenes, objects, and interactive tasks. Additionally, it includes scenes simulating low-vision conditions with gaze tracking. This allows a deeper investigation into how well models can capture individual behavior, or generalize across behaviors in varied contexts, addressing the limits of existing computer vision approaches and datasets. 

\begin{figure}
\includegraphics[width=12.5cm]{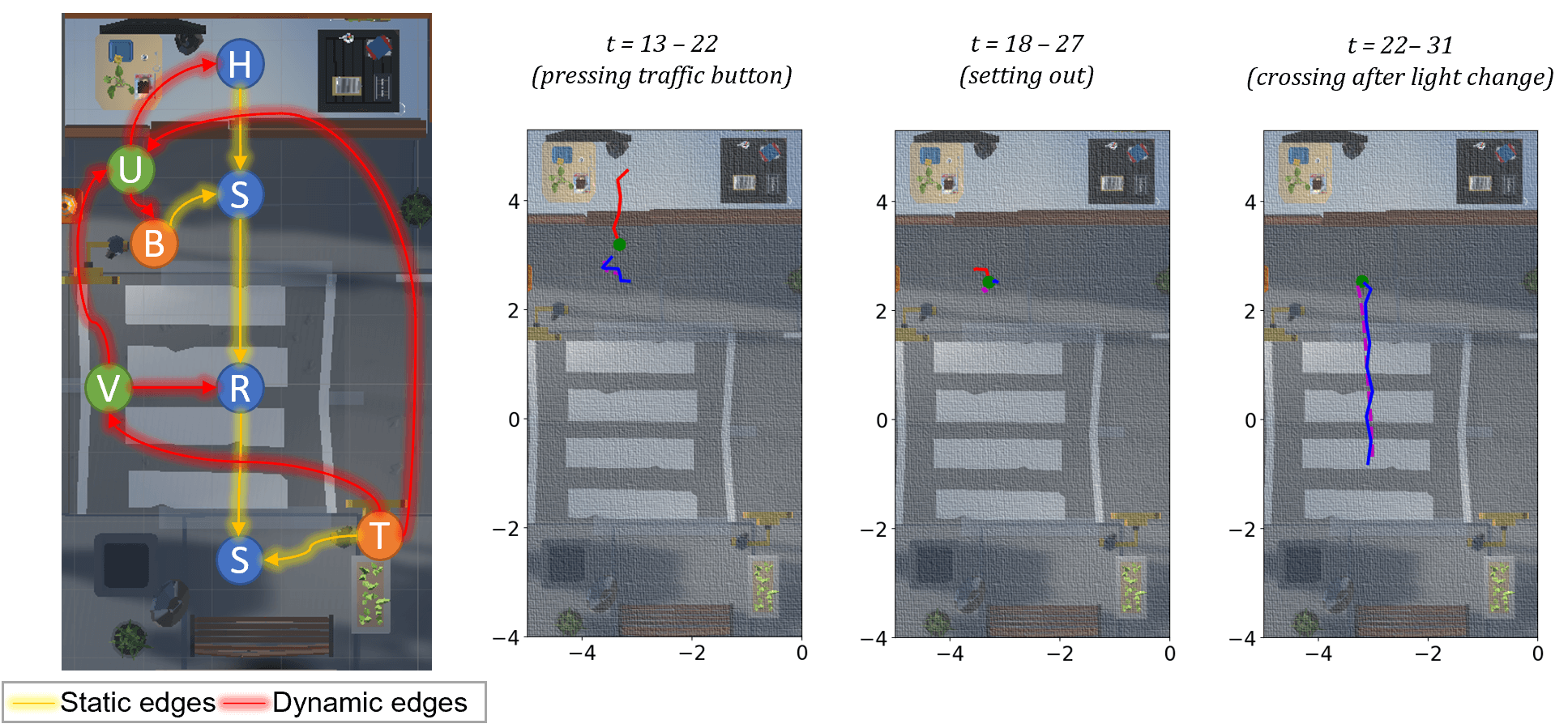}
\caption{Left: A heterogeneous graph representing a pedestrian crossing scene, with nodes for different locations (H: Home, S: Sidewalks, R: Road) and objects (U: User, V: Vehicle, T: Traffic Lights, B: Button). Right: Three scenarios of road crossing are depicted. In each scenario, the red line represents the input trajectory ending at the green dot. The blue line indicates the ground truth future motion, while the magenta line shows the predicted future motion using DiVR.}
\label{fig:teaser}
\end{figure}

Recognizing these limitations, our research introduces DiVR, a novel model that uses heterogeneous graph representations\cite{CARRASCOLIMEROS2023104405} to effectively capture the dynamic nature of human environments, as shown in Figure \ref{fig:teaser}. We leverage the annotations in the CREATTIVE3D dataset to test and refine DiVR for human motion prediction. By incorporating both static and dynamic variables, DiVR captures interactions and environmental factors, making it capable of generalizing across varying users, task complexities, and scene layouts, a key contribution to this work. Such versatility is crucial to applications like autonomous driving where the accuracy of predicting pedestrian movements can significantly impact safety and operational efficiency. We discuss DiVR’s adaptability to diverse scenarios including those involving simulated low-vision conditions -- vision loss that cannot be fully corrected -- and complex navigation tasks.

Our contributions are thus threefold:

\smallskip

\noindent 1. Propose DiVR, a novel approach using dynamic heterogeneous graphs that capture both the static and dynamic scene context to improve human motion prediction.\\
\noindent 2. Evaluate graph context to improve adaptability to low-vision and task complexity.\\
\noindent 3. Explore how scene graph context helps the model generalize better across different user groups, tasks, and scenes. 

\smallskip

\section{Related Work}
\label{sec:related_work}

\paragraph{\textbf{Human trajectory prediction.}}

Human trajectory prediction is a computer vision task, originally developed using surveillance video data from datasets like ETH and UCY. It is formulated as a sequence-to-sequence prediction task: given past position data to predict future sequences. Up until recently, Recurrent Neural Networks (RNNs) were dominantly used for trajectory prediction, including human-robot interaction scenarios \cite{zhang2020recurrent}. Corona et al. included context through simple graph representations \cite{corona2020context}, and Rond\'{o}n et al. \cite{rondon2021track} with dedicated Long Short-Term Memory (LSTM) units fed with estimated saliency maps to modulate the importance of context over inertia depending on the prediction horizon. Transformer models \cite{vaswani2017attention} marked a significant shift, introducing self-attention that captures temporal relations more effectively than RNNs, and setting new standards for motion prediction \cite{aksan2021spatio,zaier2023cross,zheng2022gimo}. However, 
transformers encounter challenges with the quadratic complexity in computational, particularly for high-dimensional inputs like images and videos, leading to significant computational costs during both training and testing phases. 
The Perceiver architecture \cite{jaegle2021perceiver} was conceived to address this by avoiding costly self-attention operations in the input space, instead restricting them to a low-dimensional latent space, with input and output dimensions handled only through cross-attention operations.

\paragraph{\textbf{Cross-modality learning.}}
Foundational cross-modal learning has been conceived to effectively fuse information from audio-visual \cite{luo2020c4av} and audio-lingual \cite{li2021ctal} tasks, and are well-adapted to motion prediction tasks. 
Cross-modal representations enable the inclusion of human behavior and environmental interaction \cite{yasar2024posetron} information.  GIMO \cite{zheng2022gimo} is representative of this and has strongly inspired our work, including motion capture, 3D point clouds, and gaze tracking in a cross-modal transformer architecture for human motion prediction. Zaier et al. \cite{zaier2023cross} introduce a cross-attention model CMAPT to fuse pedestrian trajectory data with kinematic and visual features extracted from camera surveillance videos. They employ a bi-modal transformer to capture spatio-temporal interactions. Finally, Gallo et al. \cite{gallo2024human} were the first to benchmark trajectory prediction in fully VR using the CREATTIVE3D dataset including motion, gaze, and scene point cloud data.

\paragraph{\textbf{Graph-based techniques.}}

Graph Neural Networks (GNN) have been recognized early on for their potential to model molecular structures \cite{duvenaud2015convolutional, kearnes2016molecular} and are now applied to video tasks such as action recognition \cite{gao2022classification}. In pedestrian trajectory precition, graphs can appropriately capture the interactions between people and other scene elements such as obstacles and pathways to improve accuracy \cite{zhou2020graph, yan2018spatial, sun2023exploring}. 
Graph models can involve static and dynamic scene elements, which help model continuous changes within interaction networks \cite{li2017situation}. Notably, Zhou et al. \cite{zhou2023global} introduced a hybrid static-dynamic graph approach to represent relations between pedestrians, significantly improving multi-pedestrian trajectory prediction. 
Sun et al. \cite{sun2023exploring} studied how road graphs improved trajectory prediction on the autonomous driving dataset nuScenes. Spatial Temporal GNN \cite{chen2021stgnn} effectively combines RNN mechanisms with GNNs for trajectory prediction in complex real-world scenarios such as urban and parking environments, leveraging temporal dynamics to refine predictions and align with rapidly changing settings. 
To the best of our knowledge, our approach is the first to leverage heterogeneous graph representations to investigate trajectory prediction of real humans in fully interactive virtual environments, incorporating simulated low-vision conditions. Drawing inspiration from methodologies in autonomous driving, such as those by Carrasco et al. \cite{CARRASCOLIMEROS2023104405} and Zipfl et al. \cite{zipfl2023relation}, we incorporate temporal heterogeneous graphs to effectively capture and model the dynamic interactions and temporal evolution within urban pedestrian environments.

\section{Approach}
\label{sec:approach}

Our work addresses human trajectory prediction on the CREATTIVE3D dataset, modeled on the 2D head position from past positions and context data. The human model comprises, at a given time \(t\) (in frames), the head position \(\mathbf{p}_t \in \mathbf{R}^2\) in meters, representing the user's absolute position within a 10 by 4 meters tracked space. The problem involves predicting a full motion sequence over a future horizon \(H\), defined as \(\mathbf{M}_{t+1:t+H} = \{(\hat{\mathbf{p}}_{t+1}), \ldots, (\hat{\mathbf{p}}_{t+H})\}\) from a given time \(t\). We use 3 seconds of past motion data sampled at 2 frames per second to predict the trajectory for the subsequent 5 seconds. 

Here, we present the graph representation for scene context, and the DiVR model which decodes this representation and other kinematic information into coordinate prediction.

\subsection{Context Representation}

While road graph representations are usually obtained from pre-extraction steps for autonomous driving datasets, the CREATTIVE3D dataset offers detailed annotations for entities like pedestrians, vehicles, traffic lights, and movable objects, as well as navigable spaces: house, sidewalks, roads, and crossings. Using these detailed annotations, we generate scene graphs where nodes represent individuals, entities, and spaces, and the edges represent their relations such as proximity and interactions.

To prepare the dataset, we normalize spatial coordinates and convert categorical attributes into numerical values using one-hot encoding. In our study, we explore two types of graph representations: heterogeneous and homogeneous.

\paragraph{\textbf{Heterogeneous Graph Representation}} We model the scene context as a directed heterogeneous graph, denoted by \(\mathcal{G} = (\mathcal{V}, \mathcal{E}, \mathcal{T}, \mathcal{R})\). Each node \(v \in \mathcal{V}\) and edge \(e \in \mathcal{E}\) is classified by type, with \(\tau(v) : \mathcal{V} \rightarrow \mathcal{T}\) mapping nodes to node types \(\mathcal{T} = \{\text{location}, \text{pedestrian}, \text{vehicle}, \text{button}, \text{traffic light}\}\), and \(\phi(e) : \mathcal{E} \rightarrow \mathcal{R}\) mapping edges to relation types \(\mathcal{R} = \{\text{approach}, \text{adjacent}, \text{interaction}\}\). These types represent typical urban interactions, such as a pedestrian or vehicle approaching a location, adjacency between locations, and interactions between entities like pedestrians and traffic lights.

\noindent \textit{Node Representation}: At a given time \(t\), each node represents urban entity attributes within the graph. Node attributes are detailed in Table \ref{table_node}, and include a unique identifier, interaction capability, mobility, presence status, and spatial positioning. This modeling captures both dynamic and static properties of entities, essential for representing urban interactions in our trajectory prediction task.

\begin{table}[h!]
\small
\centering
\caption{Node Attributes for the heterogeneous graph}
\label{table_node}
\begin{tabular}{ll}
\toprule
\textbf{Attribute} & \textbf{Description} \\
\midrule
$id$ & Unique identifier for the node \\
$interactable \in \{0, 1\}$ & Indicates if the node can engage with other nodes \\
$localization \in \{0, 1\}$ & Indicates if the node represents a fixed location \\
$movable \in \{0, 1\}$ & Indicates if the node is dynamic \\
$color \in \{-1, 0, 1, 2\}$ & Traffic light: red (0), orange (1), green (2), or -1 if N/A \\
$presence \in \{0, 1\}$ & Indicates if the node is present in the scene \\
$x_{\text{min}}, y_{\text{min}}, x_{\text{max}}, y_{\text{max}}, x_{\text{cent}}, y_{\text{cent}}$ & Normalized bounding box and centroid coordinates \\
\bottomrule
\end{tabular}
\end{table}

\noindent \textit{Edge Representation}: Edges in the heterogeneous graph indicate connectivity and are annotated with features that describe the nature of interactions, such as active or inactive status, type of interaction (e.g., crossing, waiting), and the physical or temporal distance between entities. This attributed edge characterization allows us to distinguish, for example, a pedestrian waiting at a traffic light from one actively crossing the street, providing deeper insights into interactive dynamics. Table \ref{table_edge} outlines the edge attributes.

\begin{table}[h!]
\small
\centering
\caption{Edge Attributes for the heterogeneous graph}
\label{table_edge}
\begin{tabular}{ll}
\toprule
\textbf{Attribute} & \textbf{Description} \\
\midrule
$id$ & Unique identifier for the edge \\
$location\_type \in \{0, 1\}$ & Indicates the position of the node \\
$dynamic\_type \in \{0, 1\}$ & Indicates if the interaction changes over time \\
$adjacent\_type \in \{0, 1\}$ & Indicates spatial adjacency \\
$interaction\_type \in \{0, 1\}$ & Indicates an interaction \\
$active \in \{0, 1\}$ & Indicates if the interaction is currently taking place \\
$distance$ & Normalized spatial distance between the connected nodes \\
\bottomrule
\end{tabular}
\end{table}

\noindent \textit{Graph Embedding ($f_c$)} To extract graph embeddings from the temporal heterogeneous graphs, we employ a Temporal Graph Convolutional Network (TemporalGCN) \cite{Gilmer2017NNConv, Simon2017ECC}, as illustrated in Fig. \ref{fig:model}. This model uses edge-conditioned convolution operations to transform edge features into dynamic weights for graph convolution layers. This adaptation allows the model to effectively differentiate and represent diverse interaction patterns within the graph. Next, global mean pooling aggregates these features across all nodes, creating a graph-level representation for each timestamp. The temporal embedding obtained from these timestamps are then concatenated, providing a representation of the dynamic graph context. Finally, a linear transformation is applied to the concatenated embedding, resulting in the final output graph embedding that integrates both spatial and temporal information.

\paragraph{\textbf{Homogeneous Graph Representation}} We model the scene context as a homogenous graph denoted by \(\mathcal{G}_{\text{h}} = (\mathcal{V}_{\text{h}}, \mathcal{E}_{\text{h}})\), where all nodes \(v \in \mathcal{V}_{\text{h}}\) and edges \(e \in \mathcal{E}_{\text{h}}\) are treated uniformly without distinct types. In this representation, nodes do not differentiate between specific types of entities, such as pedestrians, vehicles, or traffic infrastructure. Each node is associated with attributes that encompass both the type of entity and its current state.

\subsection{The DiVR Model}

\begin{figure}
\includegraphics[width=13cm]{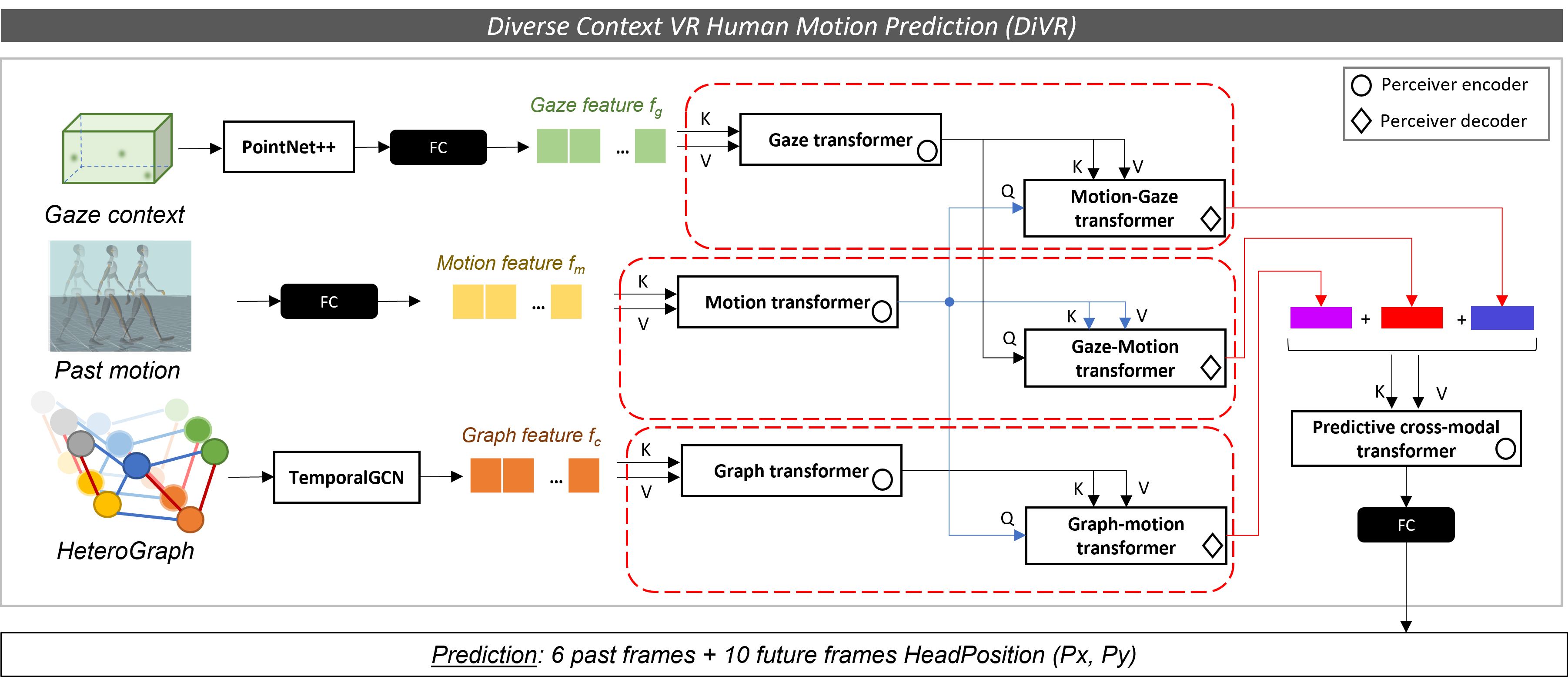}
\caption{Overview of the DiVR model}
\label{fig:model}
\end{figure}

We present the DiVR model (Figure \ref{fig:model}), a new architecture
that leverages cross-modal attention over three modalities of data: (1) gaze-interpolated scene point cloud, (2) past motion data, and (3) the human-scene interaction context through heterogeneous graph representations. Each of these modalities is processed by an individual branch with the PerceiverIO architecture. The first branch employs PointNet++ to extract and encode features from the gaze data interpolated with the scene point cloud into a latent vector \( f_{gaze} \). Concurrently, the second branch transforms raw motion data into a latent representation \( f_{motion} \). The third branch utilizes the TemporalGCN introduced in the previous section to handle temporal heterogeneous graphs, producing a latent graph vector \( f_{context} \).

At the core of DiVR's architecture is a cross-modal attention mechanism that fuses \( f_{motion} \) and \( f_{gaze} \), enhancing the model's sensitivity to the interplay between gaze direction and movements. This mechanism also integrates \( f_{context} \), combining motion, gaze, and environmental graph data via a predictive cross-modal transformer. As shown in Sec. \ref{sec:results}, this fusion of diverse modalities, along with a structured representation of the environment and interactions using heterogeneous graphs, substantially improves the accuracy of future trajectory predictions.

\noindent\textbf{Training.} To effectively train the motion prediction models under various conditions, the loss function $L_{\text{total}}$ integrates components to capture key aspects of prediction accuracy: \( L_{\text{total}} = L_{\text{trans}} + L_{\text{rec}} + L_{\text{des\_trans}} \). The translation loss, $L_{\text{trans}}$, represents the mean of the L1 losses between the predicted and ground-truth trajectories. The reconstruction loss, $L_{\text{rec}}$, penalizes discrepancies between the model's outputs and the inputs during the encoding phase, ensuring temporal consistency. The destination loss, $L_{\text{des\_trans}}$, specifically targets the accuracy of the final step predictions.

\section{Experimental evaluation}
\label{sec:results}
This section evaluates the DiVR model's performance in predicting human trajectories using the CREATTIVE3D dataset. We assess the model's adaptability comparing it against established benchmarks. The goal is to demonstrate DiVR's robustness and accuracy across different user conditions, scene complexities, and task variations, highlighting its reduced trajectory prediction error compared to conventional models. We aim to show that DiVR integrates diverse contextual data effectively, using advanced graph-based techniques.

\subsection{Experimental Setup}
\paragraph{\textbf{Dataset.}}We use the CREATTIVE3D dataset ~\cite{wu:hal-04429351} to analyze human interactions and navigation in VR environments, focusing on road crossing scenarios. This dataset, which records user behaviors in urban settings, is structured around six scenarios, the dataset includes tasks of varying complexity: from simple tasks (ST) requiring navigation to complex tasks (CT) involving interactions, such as activating a traffic light before crossing. These scenarios are tested under two visual conditions: Normal Vision (NV) and Simulated Low Vision (LV). The Low Vision condition specifically simulates a scotoma, a central blind spot. We split the dataset with 70\% for training, 15\% for validation, and 15\% for testing.


\paragraph{\textbf{Experiments.}} Multiple experiments are conducted to assess model performance under varying conditions. The \textbf{Context Data Evaluation} assesses the impact of different visual conditions and task complexities using the balanced dataset. The \textbf{Low Vision and Complex Task Generalization} experiment tests model adaptability through three configurations, beginning with a baseline model trained on NV and ST, and progressively including more complex tasks and varied visual conditions. The \textbf{User Generalization} examines the model's ability to generalize across different users by selecting a test group of 10 users at random from a total of 40. The \textbf{Scene Generalization} trains models on single-lane road scenarios and tests them on two-lane scenarios to evaluate generalization to different environments. Finally, the \textbf{Task Generalization} assesses models on crossing tasks and their adaptability to tasks requiring a return to the starting point.

\paragraph{\textbf{Evaluation metrics.}}
Our evaluation adopts two metrics that are endorsed and utilized in existing literature \cite{salzmann2021trajectron++, yu2020spatio, zhao2019multi, zaier2023cross}: the Average Displacement Error (ADE) and the Final Displacement Error (FDE). ADE measures the average L2 distance (in meters) between predicted and actual trajectories at each time step from \(T_{\text{obs}}+1\) to \(T_{\text{pred}}\), indicating overall prediction consistency. FDE calculates the Euclidean distance (in meters) between the predicted and actual final positions at the last time step, \(T_{\text{pred}}\), reflecting the accuracy of endpoint prediction.

\paragraph{\textbf{Implementation details.}}
Our model uses the PerceiverIO architecture for motion prediction. It consists of 6 encoding layers, each with a dimension of 256 and 8 attention heads, identical to the original PerceiverIO. The implementation was done using PyTorch on an Ubuntu server, supported by an Nvidia V100 GPU with 32 GB RAM. We trained the DiVR model end-to-end for 100 epochs with a batch size of 16. The Adam optimizer was used with an initial learning rate of \(0.0001\) and a weight decay of \(0.0001\). The learning rate was adjusted using an exponential decay method with a gamma value of \(0.99\).

\subsection{Results}

\subsubsection*{Context Data Evaluation}

\paragraph{\textbf{Baseline Models.}}
We compare three prediction models with increasing context integration on the CREATTIVE3D dataset: (1) MLP \cite{guo2023back} baseline, using only historical position data, (2) TRACK \cite{rondon2021track}, incorporating real-time gaze data for user attention, and (3) GIMO \cite{zheng2022gimo}, which includes gaze and 3D point cloud data and inspired our model.

\begin{table}[h!]
\small
\caption{Comparison of ADE and FDE on baseline models and DiVR. Lower scores indicate better performance. Each model takes 3 seconds input to predict the next 5 seconds.}
\label{table_context}
\footnotesize
\centering
\begin{tabular}{lcc}
\toprule
Architecture & ADE & FDE \\
\midrule
MLP \cite{guo2023back} & 0.854 & 1.512 \\
TRACK \cite{rondon2021track} & 0.759 & 1.247 \\
GIMO \cite{zheng2022gimo} & 0.757 & 1.240 \\
DiVR-Hom & 0.621 & 0.993 \\
DiVR-Het & \bf{0.588} & \bf{0.842} \\ 
\bottomrule
\end{tabular}
\end{table}

\paragraph{\textbf{DiVR Models.}}
As shown in Table \ref{table_context}, DiVR-Het not only outperforms the MLP baseline, reducing the ADE and FDE by 31.2\% and 44.3\%, respectively, but also demonstrates the effectiveness of integrating high-level context as compared to GIMO. This context includes elements such as scene layout, and human interactions, which significantly influence trajectory predictions. Similarly, DiVR-Hom shows significant improvement over the baseline with a 27.3\% reduction in ADE and 34.3\% in FDE. We show in Fig. \ref{fig:teaser} qualitative results of the DiVR-Het model.

\subsubsection{Low Vision and Complex Task Generalization}
\paragraph{\textbf{Baseline}}
Table \ref{table_gen_baseline_NV_ST} presents the baseline performance and generalization capabilities of various models trained under Normal Vision (NV) with Simple Tasks (ST), and their generalizability to complex tasks and low vision conditions. In the NV+ST test, the GIMO model achieves an ADE of 0.509 and an FDE of 0.745, slightly outperforming the TRACK model. Both models initially demonstrate better ADE values than DiVR.

\begin{table}[h!]
\small
\caption{Assessment of generalization. Comparison of ADE and FDE on baseline models and DiVR across different test conditions: Normal Vision + Simple Task, complex task, and low vision. Trained under normal vision and simple tasks.}
\label{table_gen_baseline_NV_ST}
\footnotesize
\centering
\begin{tabular}{lcccccc}
\toprule
 & \multicolumn{2}{c}{NV + ST} & \multicolumn{2}{c}{Complex task} & \multicolumn{2}{c}{Low Vision}\\
\cmidrule(lr){2-7} 
Arch & ADE & FDE & ADE & FDE & ADE & FDE \\
\midrule
MLP \cite{guo2023back} & 0.652 & 1.139 & 1.135 & 2.174 & 0.648 & 1.169 \\
TRACK \cite{rondon2021track} & 0.515 & 0.706 & 1.270 & 2.229 & 0.636 & 0.943 \\
GIMO \cite{zheng2022gimo} & \bf{0.509} & 0.745 & 1.135 & 2.163 & 0.630 & 0.951 \\
DiVR-Hom & 0.561 & 0.830 & 1.172 & 2.248 & 0.669 & 1.029 \\
DiVR-Het & 0.604 & \bf{0.696} & \bf{0.910} & \bf{1.661} & \bf{0.615} & \bf{0.860} \\
\bottomrule
\end{tabular}
\end{table}

The detailed results show that under complex task test, the DiVR-Het model adapts well, achieving an ADE of 0.910 and an FDE of 1.661. These results represent a 19.9\% reduction in ADE and 23.3\% in FDE compared to GIMO, and 28.3\% and 25.4\% reductions, respectively, compared to TRACK. In low vision condition tests, DiVR-Het also outperforms the baseline models, with an ADE of 0.615 and an FDE of 0.860. This reflects a 2.4\% improvement in ADE and 9.6\% in FDE over GIMO, and a 3.3\% improvement in ADE and 8.8\% in FDE over TRACK.


\paragraph{\textbf{Adaptation to Complex Tasks.}} Differently from the previous setup, where models were trained only on simple tasks, the models are now trained on both simple and complex tasks with normal vision (Diverse Task Training). Table \ref{table_combined_generalization_NV_DT_DV_ST}(a) shows the performance under these conditions, demonstrating that DiVR models benefit significantly from this training strategy. DiVR-Hom reduces its ADE from 0.592 to 0.470 and its FDE from 0.908 to 0.816, marking a 20.6\% improvement in ADE and a 10.1\% improvement in FDE, outperforming TRACK and GIMO under the same conditions. Although DiVR-Het's improvement in ADE is less pronounced than that of DiVR-Hom, it still demonstrates strong generalizability and adaptability.

\begin{table}[h!]
\small
\caption{Comparison of ADE and FDE between baseline models and DiVR across Different test conditions: Normal Vision + Simple Task, complex task, and low vision. Two train conditions: Diverse Task and Diverse Vision}
\label{table_combined_generalization_NV_DT_DV_ST}
\footnotesize
\centering
\begin{tabular}{lcccccccc}
\toprule
 & \multicolumn{4}{c}{(a) Diverse Task Training} & \multicolumn{4}{c}{(b) Diverse Vision Training} \\
\cmidrule(lr){2-5} \cmidrule(lr){6-9}
& \multicolumn{2}{c}{NV + ST} & \multicolumn{2}{c}{Complex Task} & \multicolumn{2}{c}{NV + ST} & \multicolumn{2}{c}{Low Vision} \\
\cmidrule(lr){2-3} \cmidrule(lr){4-5} \cmidrule(lr){6-7} \cmidrule(lr){8-9}
Arch & ADE & FDE & ADE & FDE & ADE & FDE & ADE & FDE \\
\midrule
MLP \cite{guo2023back} & 1.063 & 2.116 & 0.668 & 1.398 & 1.136 & 2.273 & 0.852 & 1.764 \\
TRACK \cite{rondon2021track} & 0.916 & 1.526 & 0.682 & 1.343 & 0.866 & 1.460 & 0.664 & 1.126 \\
GIMO \cite{zheng2022gimo} & 0.617 & 1.007 & 0.589 & 1.109 & 0.771 & 1.267 & 0.662 & 1.180 \\
DiVR-Hom & 0.592 & 0.908 & \bf{0.470} & \bf{0.816} & \bf{0.547} & 0.763 & \bf{0.560} & 0.791 \\
DiVR-Het & \bf{0.484} & \bf{0.694} & 0.506 & 0.852 & 0.574 & \bf{0.750} & \bf{0.560} & \bf{0.767} \\
\bottomrule
\end{tabular}
\end{table}

\paragraph{\textbf{Adaptation to low vision conditions.}}
Table \ref{table_combined_generalization_NV_DT_DV_ST}(b) shows the performance when models are trained on normal and low vision (Diverse Vision Training) conditions, targeting adaptability to low vision scenarios. In the low vision test, both DiVR-Het and DiVR-Hom achieve an ADE of 0.560, with FDEs of 0.791 and 0.767, respectively. These scores are lower than those of the TRACK and GIMO models.

\subsubsection{User, Scene, and Task Generalization}
\paragraph{}
This section analyzes the models' adaptability to user variations, scene complexity, and task challenges, as shown in Table~\ref{table_user_scene_task_gen}. For user generalization, 10 randomly selected users formed the test group, while others were used for training and validation. Scene generalization involved training on single-lane roads and testing on two-lane crossings. Task generalization assessed models trained on simple crossing tasks and tested on scenarios requiring a return to the starting point.

\begin{table}[h!]
\small
\caption{ Comparison of ADE and FDE values on (1) three architectures (GIMO, DiVR-Hom, and DiVR-Het) for User, scene, and task generalizability.}
\label{table_user_scene_task_gen}
\footnotesize
\centering
\begin{tabular}{lcccccc}
\toprule
 & \multicolumn{2}{c}{User} & \multicolumn{2}{c}{Scene} & \multicolumn{2}{c}{Task} \\
\cmidrule(lr){2-7}
Architecture & ADE & FDE & ADE & FDE & ADE & FDE \\
\midrule
GIMO \cite{zheng2022gimo}        & 0.621 & 0.994 & \bf{0.997} & 1.479 & 1.296 & \bf{2.100} \\
DiVR-Hom    & \bf{0.513} & 0.841 & 1.011 & \bf{1.469} & \bf{1.249} & 2.183 \\
DiVR-Het    & 0.559 & \bf{0.828} & 1.073 & \bf{1.469} & 1.384 & 2.605 \\
\bottomrule
\end{tabular}
\end{table}

\noindent\textbf{User Generalization.}
The models show good adaptability to user variations, with DiVR-Hom achieving the best ADE of \(0.513\), a \(17.4\%\) improvement over the baseline GIMO model. DiVR-Het also excels in FDE, with a \(16.7\%\) improvement, indicating both DiVR models effectively generalize across diverse user behaviors.
\textbf{Scene Generalization.}
Adapting to increased scene complexity from single-lane to two-lane crossings, DiVR models show increases in ADE and FDE to around 1 and 1.4, respectively, demonstrating reasonable resilience given the additional 3.5m complexity in two-lane roads.
\textbf{Task Generalization.}
This presents the most significant challenge, with FDE values around 2 across all models. However, DiVR-Hom achieves the best ADE, suggesting the model can infer some directional or trajectory characteristics despite the increased task difficulty.

\subsubsection*{Ablation Tests}
\paragraph{}
We conducted additional ablation studies on the DiVR-Het model. Table \ref{table_ablation} shows the results, focusing on the impact of graph and gaze data on trajectory prediction accuracy, measured by ADE and FDE. The study evaluates the DiVR-Het model by replacing either graph or gaze inputs with a tensor of zeros.

\paragraph{\textbf{Graph Ablation Study.}}
The graph ablation study reveals a significant decline in performance when graph data is omitted from the DiVR-Het model as shown in Table \ref{table_ablation}. Under NV+ST test, the graph-ablated model's ADE and FDE increase to 0.622 and 0.949, respectively, marking increases of 16.3\% in ADE and 34.6\% in FDE. The impact is even more pronounced under complex task conditions, with the ADE and FDE increasing to 0.944 and 1.767, jumps of 35.8\% and 45.1\%, respectively. In low-vision tests, the increases are 13.0\% in ADE and 24.2\% in FDE, showing again the usefulness of graph data for maintaining lower error rates across vision conditions.

\begin{table}[h!]
\caption{Comparison of ADE and FDE between DiVR-Het model with graph and gaze ablation. Lower scores indicate better performance. Each model takes 3 seconds input to predict the next 5 seconds.}
\label{table_ablation}
\centering
\begin{tabular}{lcccccc}
\toprule
& \multicolumn{2}{c}{NV + ST} & \multicolumn{2}{c}{Complex Task} & \multicolumn{2}{c}{Low Vision} \\
\cmidrule(lr){2-3} \cmidrule(lr){4-5} \cmidrule(lr){6-7}
Architecture & ADE & FDE & ADE & FDE & ADE & FDE \\
\midrule
DiVR-Het & \textbf{0.535} & \textbf{0.705} & \textbf{0.695} & \textbf{1.218} & \textbf{0.583} & \textbf{0.871} \\
Graph ablation & 0.622 & 0.949 & 0.944 & 1.767 & 0.659 & 1.082 \\
Gaze ablation & 0.590 & 0.802 & 0.793 & 1.391 & 0.673 & 1.008 \\
\bottomrule
\end{tabular}
\end{table}

\paragraph{\textbf{Gaze Ablation Study.}}
The gaze ablation study shows decreased performance, but with a slightly lesser impact than graph ablation, indicating that while gaze data is important, it is less critical than graph data in the DiVR-Het model. In NV+ST test, removing gaze data results in an increase in ADE and FDE to 0.590 and 0.802, up by 10.3\% and 13.8\% compared to the non-ablated model. For complex task tests, ADE and FDE rise by 14.1\% and 14.2\% to 0.793 and 1.391. In low vision conditions, ADE and FDE increase by 15.4\% and 15.7\%, demonstrating that gaze data significantly contributes to the model’s accuracy in predicting human trajectories, especially in challenging scenarios.

\subsubsection*{Qualitative Results}

\begin{figure}[h!]
\includegraphics[width=12cm]{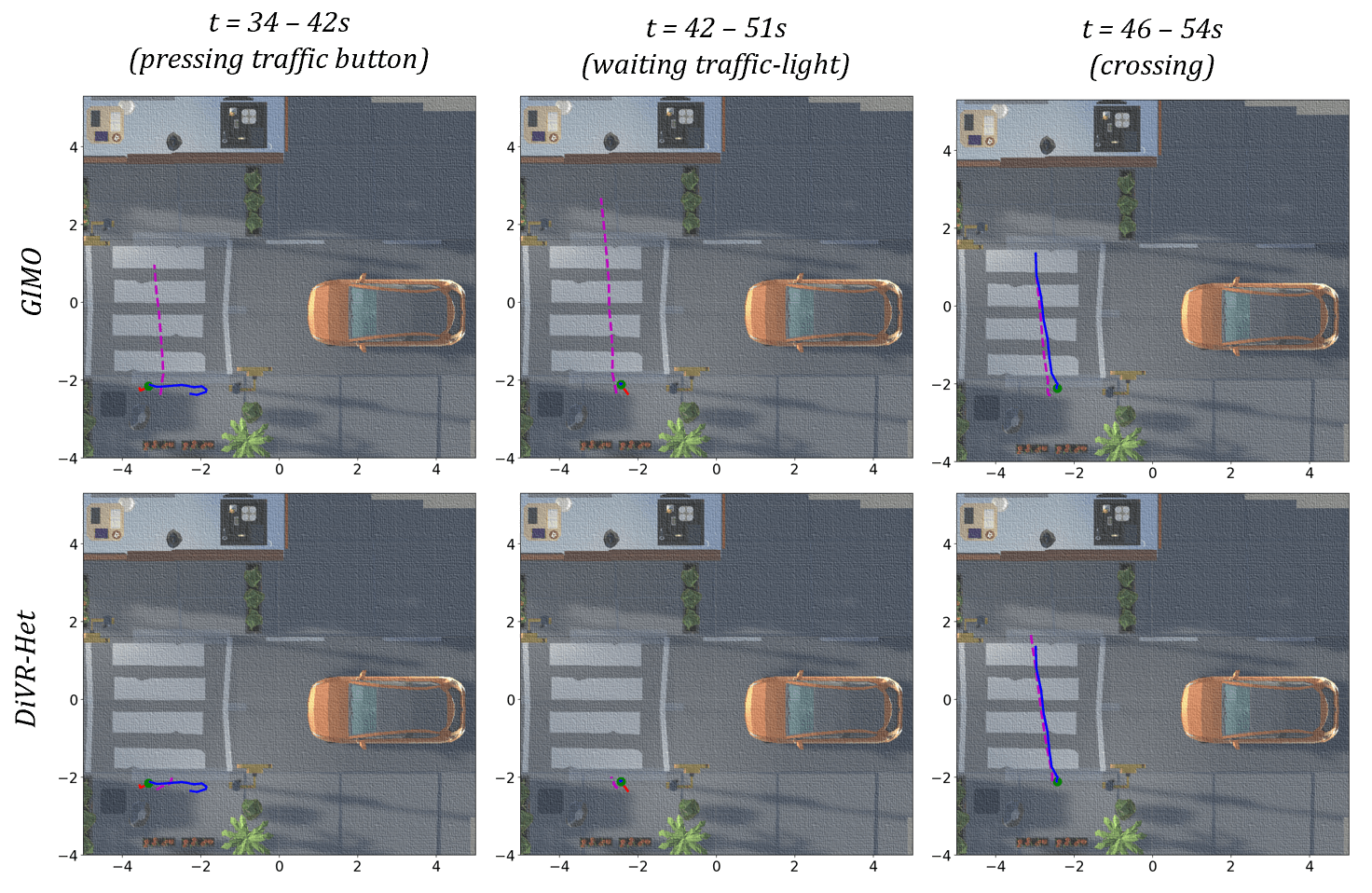}
\caption{Context evaluation: pedestrian trajectory prediction comparison between two models, GIMO (scene pointcloud) and DiVR-Het (heterogeneous graphs) in scenes sampled from CREATTIVE3D dataset. In each scene, red lines and green dots show the observed past trajectories and their endpoints, blue lines the ground truth future trajectories, and magenta lines the predicted trajectories.}
\label{fig:sup_qual}
\end{figure}

\paragraph{\textbf{Context Data Evaluation.}}
 We compare GIMO and DiVR-Het models across three urban scenes. In the first scene, "pressing Traffic Button" (left column, Fig. \ref{fig:sup_qual}), the observed trajectory (red line) shows a pedestrian disposing of trash and then interacting with a traffic button (blue line). GIMO (top row) incorrectly predicts immediate crossing, while DiVR (bottom row) accurately predicts the path toward the button, closely matching the ground truth. In the second scene, "Waiting for Traffic Light" (middle column), GIMO predicts the pedestrian begins crossing, but DiVR captures the stationary behavior, closely aligning with the ground truth. In the final scene, "crossing" (right column), DiVR’s predictions better capture the actual crossing path. These comparisons highlight DiVR’s superior performance in predicting complex pedestrian behaviors in urban environments.

\begin{figure}[h!]
\includegraphics[width=12cm]{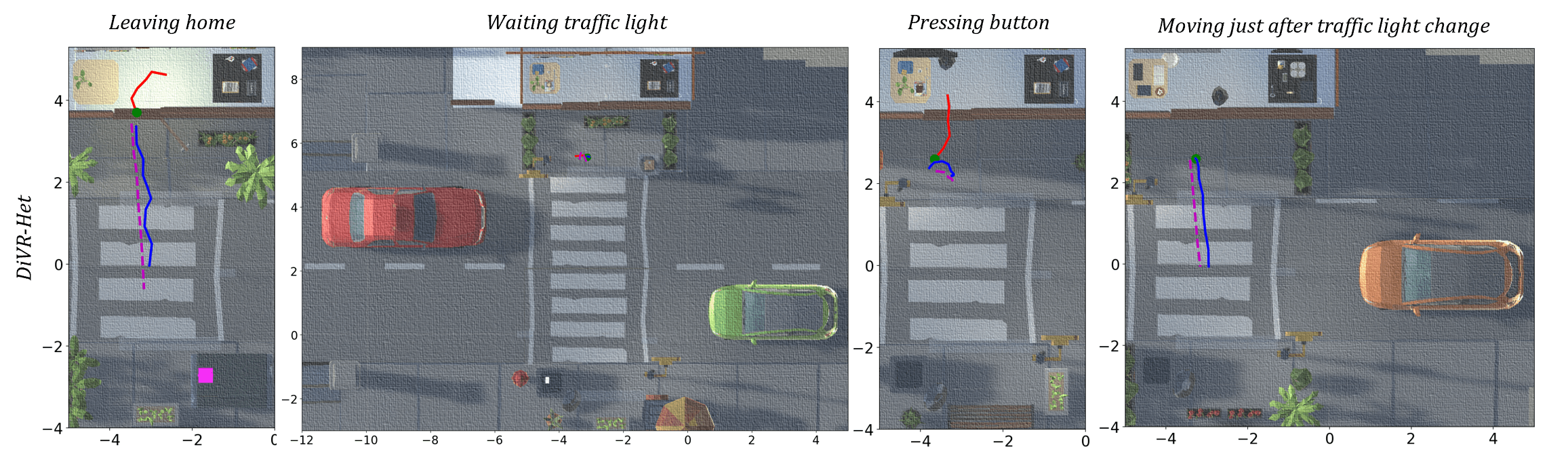}
\caption{User Generalization in CREATTIVE3D: Each subplot shows different situations (e.g. leaving home, pressing button, etc) for users that were not seen on training phase for the DiVT-het model. Predicted trajectories (magenta lines), ground truth (blue lines), observed trajectory (red lines) and green dot marks their endpoints}
\label{fig:sup_unseen_users}
\end{figure}

\paragraph*{\textbf{User Generalization}}
In the user generalization section, we visually demonstrate DiVR's adaptability across different users in the test set. Fig. \ref{fig:sup_unseen_users} illustrate scenarios where DiVR has successfully predicted diverse user movements within the CREATTIVE3D dataset, reflecting its robustness in handling variations in pedestrian behavior. Despite varied actions and intentions of users, DiVR consistently aligns closely with the ground truth, showcasing its capacity to accurately generalize across a broad spectrum of user interactions. These qualitative results complement our quantitative findings, where DiVR exhibited robust values on ADE and FDE for user generalization.

\begin{figure}[h!]
\includegraphics[width=12cm]{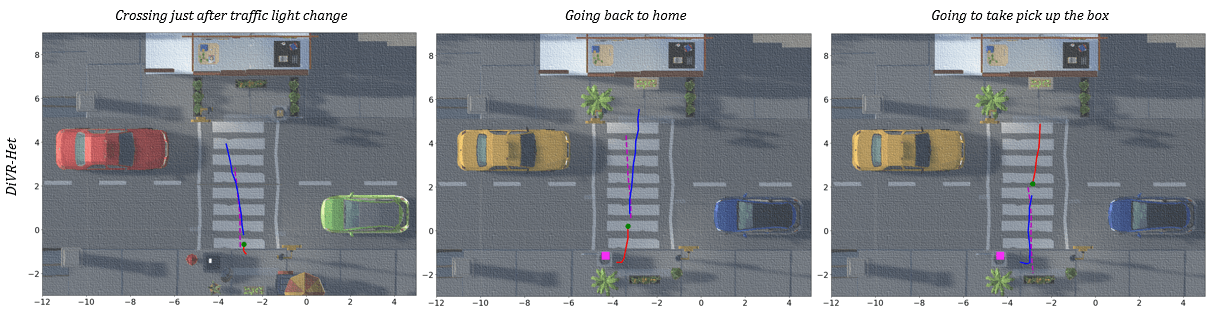}
\caption{Scene Generalization: pedestrian trajectory predictions by DiVR trained on single-lane and tested on two-lane crossing. Highlighting its capacity and limitations in adapting to complex urban layouts. Predicted trajectories (magenta lines), ground truth (blue lines), observed trajectory (red lines) and green dot marks their endpoints}
\label{fig:sup_unseen_scenes}
\end{figure}

\paragraph*{\textbf{Scene Generalization}}
For scene generalization, the visualizations focus on DiVR's performance in complex urban layouts. In Fig. \ref{fig:sup_unseen_scenes} each plot corresponds to scenarios with increased environmental complexity, from training on single-lane in training to more challenging two-lane crossing tests. While DiVR maintains a reasonable level of accuracy, the predictions reflect the heightened challenge, with slight deviations from the ground truth becoming more apparent. These instances highlight the model's limits and strengths in adapting to spatial variations and complex scene dynamics, aligning with the quantitative results show in Table \ref{table_user_scene_task_gen} that indicated a greater challenge in scene generalization compared to user behaviors.

\section{Discussion and Conclusion}
\label{sec:conclusion}
We presented DiVR, a multimodal transformer for human trajectory prediction that leverages heterogeneous graphs from rich VR contextual data. To our knowledge, this is the first work to investigate trajectory prediction of real humans in fully interactive virtual environments. Our experiments demonstrated DiVR's robustness across varied conditions, highlighted by extensive generalization tests.  A key strength of this study is the detailed evaluation across a wide range of scenarios, showcasing DiVR's effectiveness in handling complex tasks and low-vision conditions through graph-based and temporal modeling techniques. 

However, a notable limitation is the model's reliance on high-quality datasets for accurate scene graph creation, which are not widely available. To mitigate this, future work could explore data from smart city infrastructures and autonomous vehicle sensors, offering real-time traffic and pedestrian data. This would enhance the model's applicability and performance in real-world scenarios, aiding in the development of more adaptive urban traffic systems. Additionally, VR's advantages in incorporating diverse scenes and populations into training highlight its potential for important real-life applications.\\

\noindent \textbf{Acknowledgements}
This work has been partially supported by the French National Research Agency through the ANR CREATTIVE3D project ANR-21-CE33-0001 and UCA\textsuperscript{JEDI} Investissements d'Avenir ANR-15-IDEX-01 (IDEX reference center for extended reality XR\textsuperscript{2}C\textsuperscript{2}). This work was granted access to the HPC resources of IDRIS under the allocation 2024-AD011014115R1 made by GENCI.

\newpage

\bibliographystyle{splncs04}
\bibliography{main}
\end{document}